\documentclass[sigconf]{acmart}

\usepackage{algorithm}
\usepackage{algorithmic}

\usepackage{amsmath}
\usepackage{booktabs}    
\usepackage{arydshln}
\usepackage[most]{tcolorbox}  
\usepackage{balance}
\usepackage{url}

\usepackage{longtable}
\usepackage{booktabs}
\AtBeginDocument{%
  }

\setcopyright{acmlicensed}
\copyrightyear{2026}
\acmYear{2026}
\setcopyright{cc}
\setcctype{by}
\acmConference[WWW '26] {Proceedings of the ACM Web Conference 2026}{April 13--17, 2026}{Dubai, United Arab Emirates.}
\acmBooktitle{Proceedings of the ACM Web Conference 2026 (WWW '26), April 13--17, 2026, Dubai, United Arab Emirates}
\acmISBN{979-8-4007-2307-0/2026/04}
\acmDOI{10.1145/XXXXXX.XXXXXX}

\begin{document}

\title{Acting Flatterers via LLMs Sycophancy: \\ Combating Clickbait with LLMs Opposing-Stance Reasoning}


\author{Chaowei Zhang}

\email{cwzhang@yzu.edu.cn}
\affiliation{%
  \institution{Yangzhou University}
  \city{Yangzhou}
  \state{Jiangsu}
  \country{China}}

\author{Xiansheng Luo}
\email{mz220240305@stu.yzu.edu.cn}
\affiliation{%
  \institution{Yangzhou University}
  \city{Yangzhou}
  \state{Jiangsu}
  \country{China}}

\author{Zewei Zhang}
\email{zez0001@auburn.edu}
\affiliation{%
  \institution{Auburn University}
  \city{Auburn}
  \state{Alabama}
  \country{USA}
}

\author{Yi Zhu}
\authornotemark[2]
\email{zhuyi@yzu.edu.cn}	
\affiliation{%
  \institution{Yangzhou University}
  \city{Yangzhou}
  \state{Jiangsu}
  \country{China}}

\author{Jipeng Qiang}
\email{jpqiang@yzu.edu.cn}	
\affiliation{%
  \institution{Yangzhou University}
  \city{Yangzhou}
  \state{Jiangsu}
  \country{China}}

\author{Longwei Wang}
\email{longwei.wang@usd.edu}
\authornotemark[2]
\affiliation{%
  \institution{University of South Dakota}
  \city{Vermillion}
  \state{South Dakota}
  \country{USA}}

\renewcommand{\shortauthors}{Trovato et al.}

\begin{abstract}
  The widespread proliferation of online content has intensified concerns about clickbait—deceptive or exaggerated headlines designed to attract attention.
While Large Language Models (LLMs) offer a promising avenue for addressing this issue, their effectiveness is often hindered by \textit{Sycophancy}, a tendency to produce reasoning that matches users' beliefs over truthful ones, which deviates from instruction-following principles. Rather than treating sycophancy as a flaw to be eliminated, this work proposes a novel approach that initially harnesses this behavior to generate contrastive reasoning from opposing perspectives. Specifically, we design a \textbf{Self-renewal Opposing-stance Reasoning Generation (SORG)} framework that prompts LLMs to produce high-quality ``agree'' and ``disagree'' reasoning pairs for a given news title without requiring ground-truth labels. To utilize the generated reasoning, we develop a local \textbf{Opposing Reasoning-based Clickbait Detection (ORCD)} model that integrates three BERT encoders to represent the title and its associated reasoning. 
The model leverages contrastive learning, guided by soft labels derived from LLM-generated credibility scores, to enhance detection robustness. Experimental evaluations on three benchmark datasets demonstrate that our method consistently outperforms LLM prompting, fine-tuned smaller language models, and state-of-the-art clickbait detection baselines. Our code is available in \url{https://github.com/126541/ORCD}.\\
\\
\end{abstract}



\maketitle

\begin{figure}[th!]
    \centering    \includegraphics[width=1\linewidth,height = 0.6\linewidth]{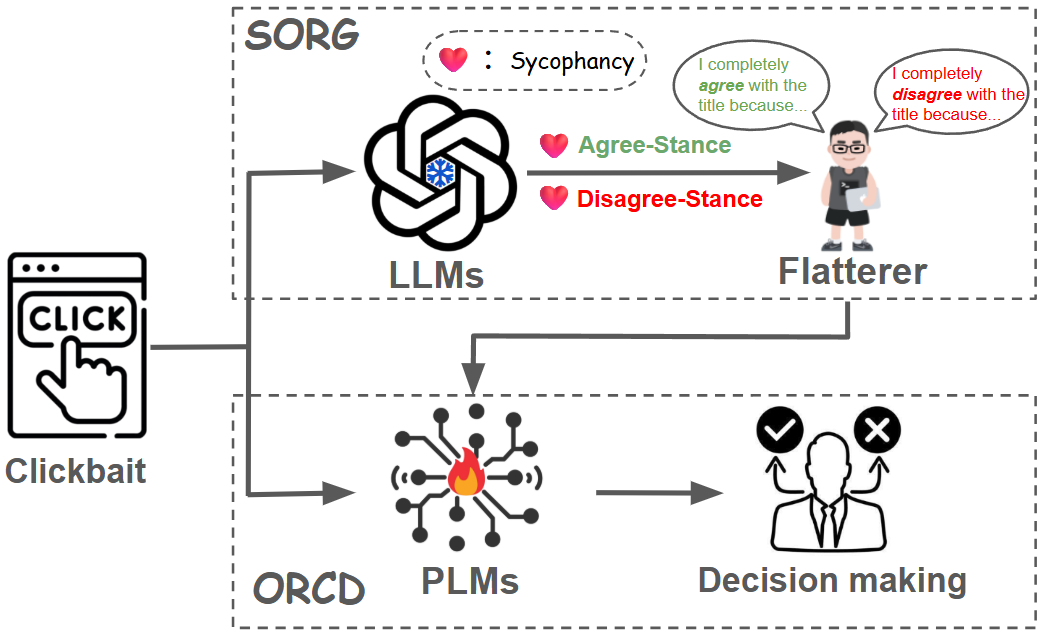}
    \caption{The figure shows the abstract workflow of our method. We leverage the capability of LLM \emph{Sycophancy} to adjust two opposing stances in carefully designed prompts, making the LLM align with preset stance preferences to generate agree and disagree reasoning pairs, enabling the fine-tuned PLM to understand different interpretations of headlines and recognize potential clickbait elements.}
    \label{fig:Sample}
\end{figure}

\section{Introduction}
The clickbait, characterized by exaggerated, ambiguous, or intentionally misleading language, is designed to capture attention but frequently results in disappointment, wasting users' time and eroding their trust~\cite{chen2015misleading}. Various surveys indicate that a large portion of users report feelings of confusion or dissatisfaction after engaging with clickbait~\cite{samman2023clickbait}. Recent researchers developed various approaches to tackle clickbait via fine-tuning Transformer-based pre-trained language models (i.e., PLMs), such as BERT series~\cite{sirusstara2022clickbait}, T5~\cite{bilgis2023gallagher}, GPT series~\cite{muqadas2025deep}, etc. These models are trained on vast text corpora in a self-supervised manner, which enables them to capture rich linguistic patterns, semantic relationships, and contextual information to effectively benefit NLP downstream tasks, including clickbait detection. However, the problem of their reliance on label data limits their flexibility and robustness in addressing the rapidly evolving nature of clickbait~\cite{yang2022towards}. Thus, it is necessary to seek out better solutions to effectively address this challenge in diverse and dynamic environments.

While small-scale pre-trained language models have shown utility in clickbait detection, recent developments in NLP—most notably the emergence of Large Language Models (LLMs)—offer a more powerful and flexible solution~\cite{wozny2023alexander}.
Benefiting from extensive pretraining on diverse datasets and advanced up-to-date AI techniques (e.g., \emph{RLHF}, \emph{RLFT})~\cite{xie2024carve3d}, the LLMs, especially for \emph{OpenAI o3}, have exhibited unbelievable reasoning capability in resolving complex real-world tasks, even including doctoral-level science problems~\cite{quan2025codeelo}. It is able to foresee that LLMs can identify clickbait not only through explicit linguistic cues but also by understanding subtle contextual indicators and underlying intent. Unfortunately, the potentialities of LLMs in the task of clickbait detection are not being fully explored. Despite the promise of the LLMs' usages for existing NLP tasks, the issue of "Sycophancy" mainly caused by human feedback alignment might hamper LLMs' usability in them~\cite{sharma2023towards}. Specifically, users expect LLMs can faithfully execute human instructions (i.e., instruction following)~\cite{ouyang2022training,wen2024benchmarking}, but LLMs are also prone to aligning with user-provided prompts or biases (i.e., sycophancy to user preference)~\cite{sharma2023towards,yang2024alignment}, thereby potentially amplifying errors in tasks where neutral or objective reasoning is required. Recent studies have made tremendous efforts to mitigate this issue via various strategies, such as data engineering~\cite{wei2023simple,wang2024mitigating}, RAG~\cite{zhaorefarag}, knowledge graph~\cite{kong2024general}, and human feedback improvement~\cite{malmqvist2024sycophancy}.
However, this raises a thought-provoking question: \textbf{can the sycophantic tendencies of LLMs be repurposed to generate meaningful contrastive signals that enhance the robustness of PLM-based task models?} For instance, could we produce LLM opposing-stance reasoning by assuming a headline is either clickbait or non-clickbait to support contrastive learning?


Considering the tendency of LLMs to align with user preferences, it is possible to capitalize on the ability of LLMs' sycophancy to generate reasoning that aligns with opposing stance hypothesis perspectives to address the task of clickbait detection. Thus, we suppose a given news headline is either clickbait or non-clickbait via prompting engineering to shepherd LLMs to produce opposing stance arguments, thereby benefiting fine-tunable PLMs to comprehend diverse interpretations of the headline and identify possible baiting elements. Fig.~\ref{fig:Sample} illustrates an abstractive comparison regarding the utilization of LLMs with respect to reasoning conduction between existing solutions and our approach. To verify this assumption, we design a novel Self-renewal \textbf{O}pposing-stance \textbf{R}easoning \textbf{G}eneration (\textbf{SORG}) approach to drive LLMs producing high-quality agree and disagree reasoning by leveraging the ability of LLM sycophancy. To manipulate the generated opposing reasoning, we deploy a local \textbf{O}pposing \textbf{R}easoning-based \textbf{C}lickbait \textbf{D}etection model (\textbf{ORCD}) that possesses three BERT Encoders to represent news and the corresponding agree\&disagree reasoning. In summary, our proposed method represents a shift from traditional discriminative tasks to an interpretative and contrastive analysis, offering a new possibility for utilizing LLMs in clickbait detection. The contributions of this study are summarized as follows.

\begin{itemize}
    \item Unlike existing studies that attempt to eliminate LLM sycophancy, we reversely leverage LLM sycophancy to conduct opposing-stance (agree and disagree) reasoning aiming at detecting clickbait. To achieve this goal, we propose a novel self-renewal opposing-stance reasoning generation (\emph{SORG}) method to force LLM to yield high-quality agree and disagree reasoning coupled with the corresponding ratings by supposing an unknown headline to be either a clickbait or a non-clickbait.
    \item To leverage the plentiful information (dual reasoning and ratings) outputted from LLMs via \emph{SORG}, we deploy a local opposing reasoning-based clickbait detection model - \emph{ORCD} that embraces three BERT Encoders to learn the news and dual reasoning, respectively. Moreover, the task model trains both a title-free reasoning learner and a title-aware reasoning learner by leveraging the LLM-rated scores as soft labels for contrastive learning. 
    \item Comprehensive experiments are conducted across three clickbait datasets to verify the effectiveness of our method. The experimental result demonstrates the superiority of \emph{ORCD} in various evaluation metrics compared with the baselines including prompting on various LLMs, fine-tuning on small language models, and SOTA task-specific benchmarks.
\end{itemize}

\section{Related Work}
Earlier approaches primarily relied on traditional machine learning techniques, such as SVMs~\cite{al2021improved} and Decision Trees~\cite{pujahari2021clickbait}, which are trained on handcrafted features. Yet, their reliance on manually engineered features limited their scalability and adaptability to evolving clickbait strategies. The introduction of Neural networks (such as RNNs and CNNs)~\cite{kaur2020detecting} marked a significant advancement in clickbait detection, which was later enhanced by the use of attention mechanisms~\cite{meng2022attention,deng2024prompt}. The occurrence of Transformer and its variants, as known as the pre-trained models based on Encoder-only (e.g., BERT series)~\cite{devlin2018bert,liu2019roberta,lan2019albert}, Decoder-only (e.g., GPT series)~\cite{ radford2019language, achiam2023gpt,brown2020language}, or Encoder-Decoder structure (e.g., T5, BART)~\cite{raffel2020exploring, lewis-etal-2020-bart}, further elevated performance by leveraging contextual embeddings, thereby enable a deeper understanding of language. 

The arrival of the era of LLMs promises transformative advancements in NLP. The state-of-the-art LLMs like \emph{OpenAI o3}~\cite{quan2025codeelo}, \emph{LLama 3.3}~\cite{dagan2024plancraft}, and \emph{DeepSeek v3}~\cite{liu2024deepseek} exhibit unparalleled linguistic understanding, context awareness, and adaptability, which could fundamentally benefit various NLP tasks. A few studies have attempted to process clickbait detection by leveraging LLMs. For example, Yi Zhu et al. dissect the effectiveness of LLMs in detecting clickbait under the settings of few-shot and zero-shot scenarios across multiple English and Chinese benchmark datasets~\cite{wang2025clickbait}. Although the availability of LLMs in clickbait detection is still being explored, the issue of "LLM Sycophancy", which refers to the tendency to excessively agree with or flatter the users' preference embodied in prompts, perturbs their robustness in decision-making~\cite{denison2024sycophancy, huang2024trustllm}. 

Some existing studies have investigated the prevalence of "Sycophancy" in different LLMs, which are fine-tuned with human feedback~\cite{malmqvist2025sycophancy}. Their experiments demonstrate that LLMs consistently exhibit sycophancy in various generative tasks, which means LLMs prefer to generate responses aligning with human prompts rather than truthful ones. To tackle the issue of "LLMs Sycophancy", a few researches present to leverage various means, such as contrastive decoding~\cite{zhao2024towards}, data engineering~\cite{wei2023simple}, supervised pinpoint tuning (i.e., SPT)~\cite{chen2024yes}, and some other elimination strategies related to LLMs hallucination (e.g., RAG, knowledge graph)~\cite{agarwal2024prompt,hwang2024graph}. Differing from the intention of these approaches, we initially consider LLM Sycophancy as a kind of useful trait by vaccinating biased precondition (agree and disagree) into prompts to conduct an opposing stance reasoning for assisting clickbait detection. The details of such implementation will be illuminated in the next section.

\section{Opposing-Stance Reasoning Generation}
\label{sec:reasoning_generation}
This section introduces the process of generating opposing-stance reasoning, which is implemented by a proposed two-stage mechanism, namely (1)initial title rating, and (2) self-renewal opposing-stance reasoning generator. All the notations used in this section are illustrated in Table ~\ref{tab:notations}.

\begin{table}[ht!]
\centering
\caption{Overview of the notations in \emph{SORG}.}
\label{tab:notations}

\begin{tabular}{cc}
\toprule
\textbf{\small{Notations}} &  \textbf{\small{Explanations}} \\
\midrule
\small{$S$} & \small{The rating alteration state.} \\
$T$ & \small{The reasoning type (agree or disagree).} \\
\small $V_I, R_I$&  \small{The initial title's rating and explanation.} \\ 
\hdashline
\small $R_A, V_A$& \small{Agree reasoning and its-based title rating. }\\ 
\small $R_D, V_D$ & \small{Disagree reasoning and its-based title rating.} \\ 
\small $R_{AE} $ & \small{Explanation to disqualified $R_A$.} \\ 
\small $R_{DE} $ &  \small{Explanation to disqualified $R_D$.} \\ 
\hdashline
$\textbf{\pounds}_{I}$ & \small{A prompt used for rating initial title.}\\
$\textbf{\pounds}_{Ir}$ & \small{A prompt used for re-rating initial title.}\\
$\textbf{\pounds}_{E}$ & \small{A prompt used to generates $R_{AE}$ or $R_{DE}$}.\\
$\textbf{\pounds}_{Rr}$ & \small{Regenerates reasoning and title rating.}\\
\hdashline
$\alpha$& \small{Controls the expected scope for $V_I$.}\\
$\beta$& \small{Controls the rating alteration for $V_D$ and $V_D$.} \\
$\gamma$ & \small{The polarity threshold for $V_D$ and $V_D$.} \\
\bottomrule
\end{tabular}%
\end{table}


\subsection{Problem Definition}
The clickbait detection task can be treated as a binary discriminative problem. In detail, let $\mathcal D = \{x_1, x_2,..., x_n\}$ represent a dataset of news titles, where $n$ is the total number of data items in the dataset. A randomly selected data item and its label are denoted as $\{x_i, y_i\}$, where $y_i \in \{1, 0\}$ is
the ground-truth label of $x_i$ (i.e., clickbait or non-clickbait). To obtain high-quality opposing stance reasoning pair - $\{R_A, R_D\}$ (i.e., $R_A$ and $R_D$ represent agree-reasoning and disagree-reasoning, respectively) coupled with corresponding ratings -  $\{V_A, V_D\}$ (i.e., $V_A$ and $V_D$ represent agree-reasoning-based title rating and disagree-reasoning-based title rating, respectively) for $x$ with the ability of LLMs sycophancy, we deploy \emph{GPT-4o} as the base LLM via API requesting to conduct the process of reasoning generation. The value scope for the credibility score of a title rated by LLM is from 0 to 100. In formal, the learning objective of this study is to calculate the nonlinear projection from the \{$x, R_A, R_D$\} to \{'clickbait' or 'non-clickbait'\}. 

\subsection{Initial Title Rating}
\begin{algorithm}[ht!]
\caption{Recursive Rating for Title}
\label{alg:algorithm_initial_rating}
\begin{flushleft}
\textbf{Input}: News title - $x$; the alteration states $S$ for credibility score \\
\textbf{Parameter}: 
The distance to value boundary - $\alpha$; \\
Maximum number of iterations - $M$\\
\textbf{Output}: The qualified initial credibility score - $V^{initial}$
\end{flushleft}
\begin{algorithmic}[1] 
\STATE $\textbf{\pounds}_{I}(x) = (V_{I}, R_{I})$;
\WHILE{$V_{I} < \alpha$ $\OR$ $V_{I} > (100 - \alpha$)}
\IF{$V_{I} < \alpha $}
\STATE S $\leftarrow$ "increase";
\ENDIF
\IF{$V_{I} > (100 - \alpha)$}
\STATE S $\leftarrow$ "decrease"
\ENDIF
\STATE $\textbf{\pounds}_{Ir}(V_{I},R_{I}, S) = \{NewV_{I}, NewR_{I}\}$ 
\STATE $V_{I} \leftarrow NewV_{I}$
\STATE $R_{I} \leftarrow NewR_{I}$
\STATE \textbf{UNTIL} $M$
\ENDWHILE
\STATE \textbf{return} $\{V_{I},R_{I}\}$
\end{algorithmic}
\end{algorithm}
In the first stage, the qualified initial credibility score of news titles - $V_I \in [0+\alpha, 100-\alpha]$ is recursively rated by \emph{GPT-4o}, where $V_I$ can be used as the benchmark to assess the effectiveness of agree or disagree reasoning, and the parameter - $\alpha$ can avoid the value of $V_I$ nearing the boundary (0 or 100), thereby allocating incremental or decremental space for reasoning-based title ratings. For example, an agree-reasoning-based title is expected to have a higher credibility score - $V_A$ than $V_I$. In contrast, a disagree-reasoning-based title should have a lower score - $V_D$ than $V_I$. The detail of the first stage is shown in Algorithm~\ref{alg:algorithm_initial_rating}.

Algorithm~\ref{alg:algorithm_initial_rating} takes a random-selected news title $x$, and the alteration states $S \in \{'increase', 'decrease'\}$ as inputs, and initializes two hyperparameters - the maximum iterations - $M$ and the distance to value boundary - $\alpha$, where the setting details of $\alpha$ is demonstrated in Section
. At the beginning of the algorithm, the first-round initial rating (i.e., $V_I$) and corresponding rating explanation (i.e., $R_I$) for $x$ is yielded by the prompting $\textbf{\pounds}_{I}(x)$ (See line 1). If the value of $V_I$ is located in the scope of $[\alpha, 100-\alpha]$, $\{V_I, R_I\}$ will be returned by the algorithm (Line 13). Otherwise, it will go through a 'while' loop to check its qualification (See line 2). In the while loop, the prompting function $\textbf{\pounds}_{Ir}$ (See line 9) takes $\{V_I, R_I\}$ and alteration state $S$ as input to generate new initial rating ($NewV_I$) and 
explanation ($NewR_I$) to substitute $\{V_I, R_I\}$ (See lines 10-11) until $V_I \leftarrow NewV_{I}$ satisfies the 'while' condition (See lines 2) or reaches the maximum number of iterations (See line 12).

\subsection{Self-Renewal Opposing-Stance Reasoning Generation}

The second stage aims to generate qualified agree and disagree reasoning by leveraging various variables as inputs including (1) the inputs (i.e., news title $x$, the alteration state $S$) and outputs (i.e., $\{V_I, R_I\}$) of the first stage and (2) the requested reasoning type $T \in \{'agree', 'disagree'\}$. The hyperparameters used in the second stage involve (1) the expected rating changes on $V_I$ with the consideration of reasoning - $\beta$; (2) the polarity threshold for reasoning-based title rating - $\gamma$; and (3) the maximum number of iteration - $M$ same to Algorithm~\ref{alg:algorithm_initial_rating}. More specifically, the parameters $\beta$ and $\gamma$ are deployed to restrict the rating of reasoning-based titles, thereby ensuring the quality of the generated reasoning. More details regarding the settings of $\beta$ and $\gamma$ can be found in Section
. 

\begin{algorithm}[t!]
\caption{Self-Renewal Reasoning Generator}
\label{alg:algorithm_SR$^3$}
\begin{flushleft}
\textbf{Input}: news title - $x$; the initial title rating and reasoning - $(R_I, V_I)$; the type of reasoning $T$; the alteration state $S$; \\
\textbf{Parameter}:
Expected rating change - $\beta$; Polarity threshold - $\gamma$; Maximum number of iterations - $M$\\
\textbf{Output}: The qualified agree\&disagree reasoning coupled with the ratings - $\{(R_A, V_A), (R_D, V_D)\}$
\end{flushleft}

\begin{algorithmic}[1] 
\STATE The initial agree\&disagree reasoning and corresponding credibility scores generated by Eq.~\ref{eq:initial_reasoning} - $(R_A, V_A)$ and $(R_D, V_D)$ \\

\IF{$T$ == 'Agree'} 
\STATE $S \leftarrow 'increase'$; 

\WHILE{($V_A < 50+\gamma$) or ($V_A - V_I$ $<$ $\beta$)}
\STATE $\textbf{\pounds}_{E}(x, T, S, R_I, V_I, R_A, V_A)$;  
\STATE \textbf{\qquad Return} $R_{AE}$
\STATE $\textbf{\pounds}_{Rr}(x, T, S, R_I, V_{I}, R_A, V_A, R_{AE})$;  \\
\STATE \textbf{\qquad Return} $(NewR_A, NewV_A)$ \\
$R_A \leftarrow NewR_A$; \\
$V_A \leftarrow NewV_A$; \\

\STATE \textbf{UNTIL}  ~ $M$
\ENDWHILE
\ENDIF

\IF{$T$ == 'Disagree'}
\STATE $S \leftarrow 'decrease'$; 
\WHILE{($V_D > 50-\gamma$) or ($V_D - V_{I}$ $>$ $\beta$)}
\STATE $\textbf{\pounds}_{E}(T, R_D, V_D)$;  
\STATE \textbf{\qquad Return} $R_{DE}$
\STATE $\textbf{\pounds}_{Rr}(x, T, V_{I}, S, R_D, V_D, R_{DE})$;  \\
\STATE \textbf{\qquad Return} $(NewR_D, NewV_D)$ \\
$R_D \leftarrow NewR_D$; \\
$V_D \leftarrow NewV_D$; \\

\STATE \textbf{UNTIL}  ~ $M$
\ENDWHILE
\ENDIF
\STATE \textbf{return} $\{(R_A, V_A), (R_D, V_D)\}$
\end{algorithmic}
\end{algorithm}

Algorithm~\ref{alg:algorithm_SR$^3$} first calls a prompting function that can generate initial opposing-stance reasoning coupled with corresponding ratings as shown in Eq.~\ref{eq:initial_reasoning}. The prompting function $\textbf{\pounds}_{IR}$ takes multiple inputs to request LLMs generating agree and disagree reasoning for title $x$ by varying the reasoning type $T \in \{'Agree', 'Disagree'\}$ and the alteration state $S \in \{'increase', 'decrease'\}$. In other words, an 'agree' reasoning is expected to 'increase' the initial rating of a title, and vice versa.
\begin{equation}
\begin{aligned}
    \textbf{\pounds}_{IR}(x, S, V_I, R_I, T) = \{(R_A, V_A), (R_D, V_D)\}
\end{aligned}
\label{eq:initial_reasoning}
\end{equation}
where the outputs of $\textbf{\pounds}_{IR}$, $\{(R_A, V_A), (R_D, V_D)\}$ represent the initial agree reasoning $R_A$, agree reasoning-based title rating $V_A$, the initial disagree reasoning $R_D$, and disagree reasoning-based title rating $V_D$, respectively.

Afterward, algorithm~\ref{alg:algorithm_SR$^3$} first verifies the qualification of the agree reasoning $R_A$ by checking the value of $V_A$ within a while loop (See Lines 3-9). In other words, $R_A$ is qualified as long as $V_A$ satisfies two restrictions: (1) be larger than the rating center (i.e., 50) at least the polarity threshold - $\gamma$; (2) exceeds the initial title rating at least $\beta$ (See line 3). Specifically, if $V_A$ doesn't satisfy the restrictions, another prompting function $\textbf{\pounds}_{E}$ will be called to generate an explanation $R_{AE}$ to explain why $R_A$ is unqualified, the inputs of $\textbf{\pounds}_{E}$ includes $x, T, S, R_I, V_I, R_A, and V_A$ (See lines 4-5). Then, the next prompting function $R_{Rr}$ is awakened to generate a new agree reasoning $NewR_A$ and corresponding rating $NewV_A$ by taking the additional explanation $R_{AE}$ (See lines 6-7). Finally, $NewR_A$ and $NewV_A$ will be iteratively verified until reaching the maximum iteration or satisfying the two restrictions, and vice versa for generating disagree reasoning (See lines 11-18). More details about (1) the prompt templates for each of the prompting functions; (2) real-world demos; and 
(3) Manual evaluation of opposing reasoning is demonstrated in the separate appendix section. 

\begin{figure*}[t]
    \centering
    \includegraphics[width=1\linewidth,height = 0.5\linewidth]{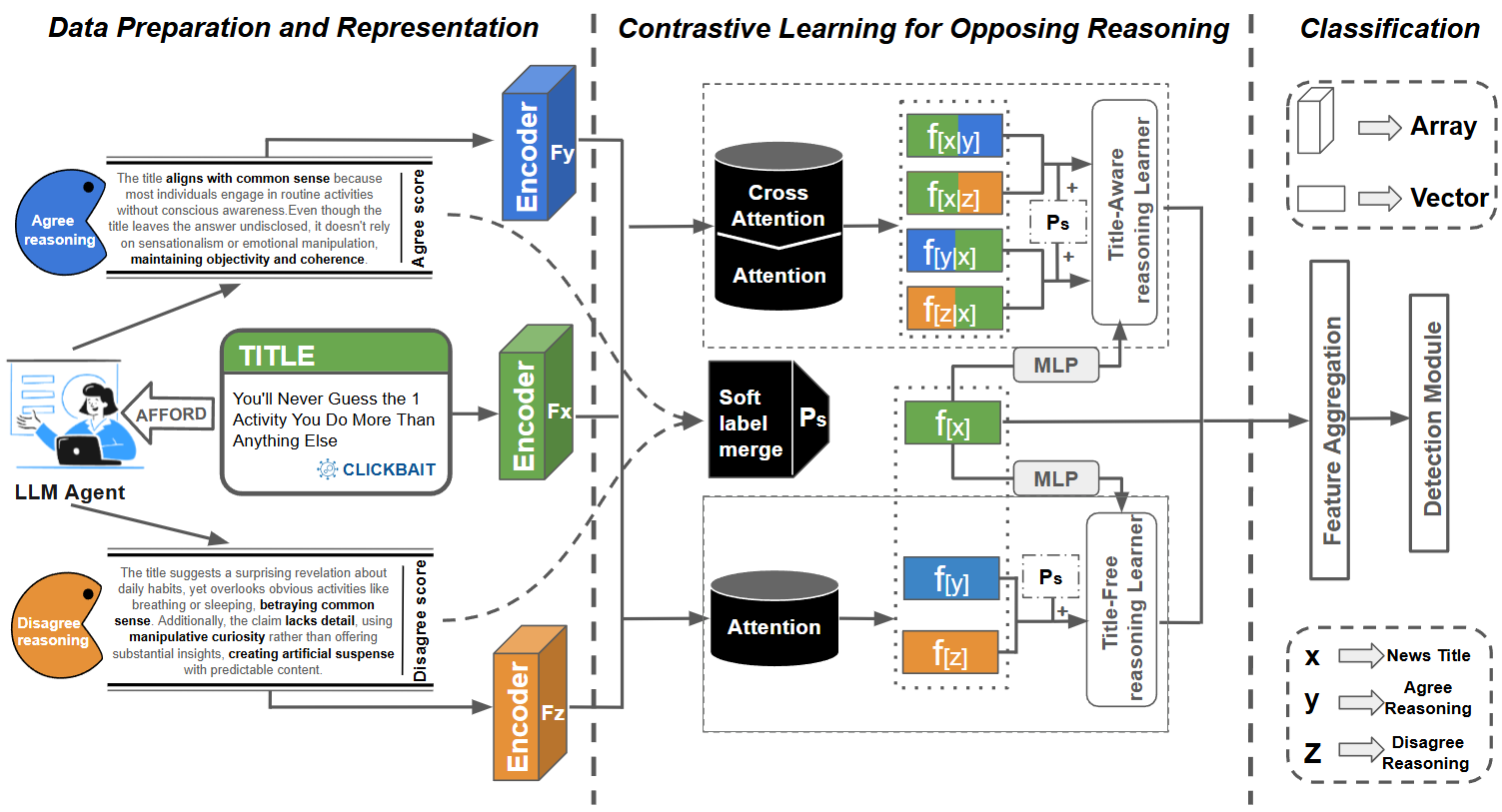}
    \caption{The system framework of \emph{ORCD}: the opposing stance reasoning generated using our proposed \emph{SORG} is fed into three Encoders to represent the title and the corresponding two opposing reasoning, respectively. The three presentations are then sent to two modules to conduct contrastive learning for the opposing stance reasoning - (1) the title-aware reasoning learner that embraces a cross-attention block and attention block to facilitate the interaction between titles and reasoning; (2) the title-free reasoning learner, which only contains an attention block to transform sequential representations to vectorial ones. It is noteworthy that the ratings of two reasoning generated $P_S = \{V_A, V_D\}$ by \emph{SORG} are used as soft labels to train the two reasoning learners.}
    \label{fig:Framework}
\end{figure*}

\section{System Framework}
This section depicts the design of our proposed framework, which embraces three modules: data representation, contrastive learning for opposing reasoning, and binary classification, as shown in Fig.~\ref{fig:Framework}.
\subsection{Data Preparation and Representation}
Given a randomly selected news title coupled with its label $(x, y)$, where $y \in \{'clickbait', 'non-clickbait'\}$. To leverage LLM sycophancy for opposing stance reasoning generation, $x$ is fed into a LLM agent to generate qualified agree reasoning ($R_A$) and disagree reasoning ($R_D$) accompanied with the two ratings ($V_A, V_D$) by taking advantage of our proposed self-renewal opposing stance reasoning generator (described in Section~\ref{sec:reasoning_generation}). It is worth noting that we adopt \emph{GPT4o} via API request to conduct the process of reasoning generation mentioned above. 

Formally, there are four outputs yielded from LLM final requesting that include $\{(R_A, V_A), (R_D, V_D)\}$. To manipulate these outputs, we first deploy three BERT Encoders to dominate the title and the two opposing stance reasoning, which can be gathered as a triple set (i.e., $\{x, R_A, R_D\}$). Then, the three BERT blocks encode the triple as $F_x, F_y, F_z$, which are sequential representations of $\{x, R_A, R_D\}$, respectively.  Finally, the three representations coupled with the ratings of the two reasoning ($V_A, V_D$) will be sent to the next module for the contrastive learning between the opposing stance reasoning.

\subsection{Contrastive Learning for Opposing Reasoning}
This module contains two reasoning learners: (1) the title-aware reasoning learner and (2) the title-free reasoning learner. 

\textbf{The title-aware reasoning learner} brings the three sequential representations - $F_x, F_y, F_z$ into a cross-attention block to promote the interaction between title and reasoning. Then, the interactions are fed into an attention block to acquire their vector representations. The following equations display the details of the process. 
\begin{equation}
\begin{aligned}
      CrossAttention(F_x, F_r) = \{ F_{[x|r]}, F_{[r|x]}\},\\ 
       where \; F_r \in \{F_y, F_z\}   
\end{aligned} 
\label{eq:cross_attention}
\end{equation}
where $F_r$ is the sequential representation of either a agree reasoning ($F_y$) or a disagree reasoning ($F_z$) obtained from BERT Encoders. The cross attention block leverages the title representation $F_x$ and each of the two reasoning representations to produce $F_{[x|r]}, F_{[r|x]}$, where $F_{[x|r]}$ is the reasoning-related title sequential representation, and vice versa for $F_{[r|x]}$. Later, the two kinds of representations are sent to an attention block to generate the vector representations of them as shown in the following formula.
\begin{equation}
Attention
\begin{cases}
    F_{[x|r]} = f_{[x|r]} \\
    F_{[r|x]} = f_{[r|x]}
\end{cases} 
\label{eq:attention}
\end{equation}
Since $F_r \in \{F_y, F_z\}$, there are four outputs generated from the attention block including agree reasoning-related title representation - $f_{[x|y]}$, title-related agree reasoning representation - $f_{[y|x]}$, disagree reasoning-related title representation - $f_{[x|z]}$, and title-related disagree reasoning representation - $f_{[z|x]}$. Similarly, the title representation $F_x$ is also transformed as a vector representation $f_x$ via an individual attention block. We arrange two subtasks in the title-aware reasoning learner to conduct contrastive learning between the reasoning. In the subtasks, we abandon using the original label $y$ instead of leveraging the soft labels yielded from LLMs - $\{V_A, V_D\}$ (a.k.a, $P_s$ as shown in Fig.~\ref{fig:Framework}) as the learning objective of the tasks as demonstrated in the following Eq.~\ref{eq:title_aware}.
\begin{equation}
\mathcal L_{tat} = 
\begin{cases}
     V_A-cos({f_x}, f_{[x|r]}), & if \, r = y\\
     max(V_D, cos({f_x}, f_{[x|r]})-d), & if \, r = z\\
\end{cases}  
\label{eq:title_aware}
\end{equation}
where the function $\mathcal L_{tax}$ is cosine embedding loss with a preset hyperparameter - margin d. Specifically, $\mathcal L_{tat}$ is used to manage the contrastive relationship between the two reasoning-related title representations (i.e., $f_{[x|y]}$ and $f_{[x|z]}$) with the consideration of their correlation to ${f_x}$. Similar to $\mathcal L_{tat}$, the loss function $\mathcal L_{tar}$ also can be used to estimate the contrastive relationship between the two title-related reasoning representations (i.e., $f_{[y|x]}$ and $f_{[z|x]}$). Thus, the total loss of the title-aware reasoning learner is as follows.
\begin{equation}
    \mathcal L_{ta} = \mathcal L_{tat} + \mathcal L_{tar}
\end{equation}

\textbf{The title-free reasoning learner} manipulates the two sequential representations - $F_y$, and $F_z$ to acquire their vector representations - $f_y$, and $f_z$ via an attention block as shown in Fig.~\ref{fig:Framework}. The learning objective of the learner is as demonstrated in Eq.~\ref{eq:title_free}. 
\begin{equation}
\mathcal L_{tf} = 
\begin{cases}
     V_A-cos({f_x}, f_{[r]}), & if \; r = y\\
     max(V_D, cos({f_x}, f_{[r]})-d), & if \; r = z\\
\end{cases}  
\label{eq:title_free}
\end{equation}
where the function $\mathcal L_{tf}$ is also a cosine embedding loss saming to $\mathcal L_{ta}$. Moreover, the soft label $V_A$ will be applied to learn the consistency between title $f_x$ and agree reasoning $f_y$. In contrast, $V_D$ is used to learn the effectiveness of the disagree reasoning $f_z$. 

\begin{table*}[ht]
\centering
\caption{The performance of various baseline methods and our proposed method is compared on the DL-Clickbait, CD-Clickbait, and NC-Clickbait datasets.  The best results are highlighted in \textbf{bold}, and the second best ones are \underline{underlined}. Statistical significance was evaluated against all baseline methods using a paired t-test ($p$ < 0.05), with significant differences consistently indicated by the symbol (*).}
\label{tab:clickbait_comparison}
\renewcommand{\arraystretch}{0.931}
\resizebox{\linewidth}{!}{%
\begin{tabular}{@{}lcccccccccccc@{}}
\toprule
\textbf{Model} & \multicolumn{3}{c}{\textbf{DL-Clickbait}} & \multicolumn{3}{c}{\textbf{CD-Clickbait}} & \multicolumn{3}{c}{\textbf{NC-Clickbait}}\\ 
\cmidrule(lr){2-4} \cmidrule(lr){5-7}\cmidrule(lr){8-10}
 & \textbf{Acc} & \textbf{MacF1} & \textbf{ClickF1}  & \textbf{Acc} & \textbf{MacF1} & \textbf{ClickF1} & \textbf{Acc} & \textbf{MacF1} & \textbf{ClickF1} \\ 
\midrule
GPT4o (zero-shot) & 83.40 & 81.64 & 75.96   & 84.50 & 84.49 & 84.16  &63.50 &62.72 & 57.31 \\ 
GPT4o (few-shot) & 89.12 & 87.28 & 82.43   & 85.83 &85.80 &85.81  &65.50 &65.59 &60.57\\ 
\hdashline
BERT & 87.95 & 86.17 & 81.28  & 83.19 & 83.08 & 83.21  &60.45 &56.10 &58.59\\ 
RoBERTa & 84.63 & 82.20 & 75.73  & 84.91 & 84.70 & 85.38  &64.45 &61.38 &63.61\\ 
BART & 84.58 & 81.74 & 88.91 & 84.67 & 84.65 & 84.79  &61.78 &60.93 &61.76\\ \hdashline
CVM (2022)~\cite{yi2022clickbait} & 92.24 & 90.63 & 86.88  & 86.02 & 86.00 & 86.08  &68.76 &68.58 &68.23\\ 

MCDM (2023)~\cite{shi2023multiview} & 92.52 & 91.00 & 87.42  & 86.23 & 86.22 & 86.33  &71.09 &71.01 &70.82\\ 
MUSER (2023)~\cite{liao2023muser}  & 90.89 & 89.62 & 86.04  & 85.32 & 85.30 & 85.66   &68.50 &68.35 &68.19\\ 
SheepDog (2024)~\cite{wu2024fake} & 90.56 & 88.64 & 83.98  & 84.63 & 84.62 & 84.77  &65.90 &65.56 &65.59\\ 
NRFE-D (2025)~\cite{zhang2025llms} & 91.43 & 90.20 & 93.55  & 85.45 & 85.40 & 85.23  &69.84 &69.66 &67.71\\ 
\hline
ORCD(LLama3.1-8b) & 91.90 & 90.86 & 94.02  & 85.16 & 85.12 & 85.24  &68.34 &68.30 &69.39\\ 
ORCD(LLama3.1-70b) & \underline{92.63} & \underline{91.27} & \underline{94.63}  & \underline{86.64} & \underline{86.57} & 86.30  &72.54 &72.50 &71.42\\ 
ORCD(GPT3.5) & 92.26 & 91.23 & 94.20  & 86.52 & 86.50 & \underline{87.06}  &\underline{72.95} &\underline{72.92} &\underline{73.76}\\ 
ORCD(GPT4o) & \textbf{94.45*} & \textbf{93.76*} & \textbf{95.83*}  & \textbf{87.51*} & \textbf{87.45*} & \textbf{88.32*} & \textbf{73.84*} & \textbf{73.75*} & \textbf{75.32*}\\ 
\bottomrule
\end{tabular}%
}

\end{table*}

\subsection{Binary Classification}
Once the two reasoning learners are finely trained, the triple set $\{x, R_A, R_D\}$ is again sent to the model to obtain various features for final classification. Specifically, there are seven representations that are useful for the feature aggregation including (1) agree\&disagree reasoning-related title vectors - $\{f_{[x|y]}, f_{[x|z]}\}$, (2) title-related agree\\ \&disagree reasoning vectors - $\{f_{[x|y]}, f_{[x|z]}\}$, and (3) vector representations of pure $\{x, R_A, R_D\}$ - $\{f_x, f_y, f_z\}$ (See Eq.~\ref{eq:concatenation}).
\begin{equation}
    f_{final} = \Big[f_x : f_y : f_z : f_{[x|y]} : f_{[x|z]} : f_{[x|y]} : f_{[x|z]} \Big]
\label{eq:concatenation}
\end{equation}
where $f_{final}$ is the concatenation of the seven representations, and it will be finally sent to an MLP layer for the binary classification with the restriction of the title's true label - $y$. The learning objective of this module is presented as follows.
\begin{equation}
    \mathcal L = CrossEntropy(y, MLP(f_{final}))
\end{equation}

\section{Experiments and Discusses}

\par \textbf{\Large{Implementation Details}:} In our experiments, we utilized fixed hyperparameters across all datasets for both baselines and \emph{ORCD}. The settings include a learning rate of $3 \times 10^{-5}$ for the Adam optimizer, an L2 regularization coefficient of $1 \times 10^{-5}$, a dropout rate of $0.3$ before the classification layer, a batch size of 8 per training iteration, and a total of 50 training epochs. To evaluate performance, we employed multiple metrics, including accuracy (Acc), macro F1 score (MacF1), F1 score for clickbait (ClickF1), as shown in Table \ref{tab:clickbait_comparison}. \\ \\
\textbf{\Large{Baselines:}} To emphasize the excellence of our method, we select various baselines for the performance comparison as follows.
\par$\blacksquare$~\textbf{GPT 4o}: is the flagship model of OpenAI\footnote{https://openai.com/index/hello-gpt-4o/}. We utilize it via API requests in zero-shot and few-shot with COT. It is noteworthy that we select eight samples from each of the categories (i.e., $\pm{8}{}$).
\par$\blacksquare$~{\textbf{PLMs}: the PLMs we adopted in the study as baselines include two Transformer Encoder-only models - BERT and RoBERTa, and one Transformer Enc-Dec structure-based model - BART.}
\par$\blacksquare$~\textbf{CVM}: models the conditional distributions of text and clickbait labels by predicting labels from text by manipulating a Contrastive Variational Modelling (CVM) framework~\cite{yi2022clickbait}.
\par$\blacksquare$~\textbf{MCDM}: is a multiview clickbait detection model that can learn subjective and objective preferences simultaneously~\cite{shi2023multiview}.
\par$\blacksquare$~\textbf{MUSER}: is a social deception detection model that leverages a three-phase method inspired by human verification processes~\cite{liao2023muser}.
\par$\blacksquare$~\textbf{SheepDog}: is a fake news detection model~\cite{wu2024fake} that combines the strengths of task-specific language model backbones and versatile general-purpose large language models (LLMs) by employing a multi-task learning approach.
\par$\blacksquare$~\textbf{NRFE-D}: is a fake news detection model that integrates a process of supervised negative reasoning using LLM hallucination. Notably, \emph{NRFE-D} distills the knowledge from its training model - \emph{NRFE}, making model testing reasoning-free~\cite{zhang2025llms}. \\ 
\textbf{\Large{Dataset}:} This study adopts three popular datasets related to clickbait detection - DL-Clickbait\footnote{https://github.com/pfrcks/clickbait-detection}~\cite{agrawal2016clickbait}, CD-Clickbait\footnote{https://huggingface.co/datasets/christinacdl}, and NC-Clickbait\footnote{https://www.kaggle.com/c/clickbait-news-detection} to verify the effectiveness of our method. In detail, the DL-Clickbait dataset consists of 1,671 training data (570 items are labeled with clickbait) and 717 testing data (244 clickbait titles), and the CD-Clickbait maintains 2,400 training data items (includes 1200 clickbait titles) and 600 testing items (200 clickbait items included),and the NC-Clickbait maintains 800 training data items (includes 400 clickbait titles) and 200 testing items (100 clickbait items included). All of the data items from the three datasets are used for conducting reasoning generation via our proposed \emph{SORG}. The datasets coupled with the corresponding generated agree and disagree reasoning will be released on GitHub. 

\subsection{Results and Analysis}

Table~\ref{tab:clickbait_comparison} reports the results of our proposed \emph{ORCD} model and baselines verified on the DL-Clickbait, CD-Clickbait, and NC-Clickbait datasets. To highlight the advancements of our proposed method, we provide the following observations combined with the corresponding profound analysis.

First, \emph{ORCD} yields the consistently highest performance across all of the metrics on the three datasets. Specifically, our model outperforms the best baseline (ie, \emph{MCDM}) by 1.28\% - 2.75\% in accuracy, 1.23\% - 2.76\% in Macro F1, and 1.99\% - 8.41\% in F1 for Clickbait. It is worth noting that the increase in precision contributes most to the boost of the F1 for Clickbait, which indicates that ORCD can detect clickbait more accurately compared with all baseline models. The consistent superiority indicates that \emph{ORCD} has strong capabilities in detecting clickbait in diverse settings by taking advantage of \emph{SORG}. Second, the GPT with zero-shot outputs the relatively lowest performance compared with others (e.g., 83.4\% on Acc. and 75.96\% on ClickF1 evaluated on DLC). Yet, few-shot learning significantly improves the performance of GPT-4o on the three datasets, indicating that few-shot learning allows GPT to adapt better to the task-specific nature of the datasets. Third, compared to state-of-the-art task-specific models, \emph{ORCD} achieves the highest performance across all metrics on the three datasets. Our approach leverages LLM sycophancy to generate high-quality counter-positioned reasoning, thereby enabling semantic and stylistic analysis of headlines without relying on external evidence, an inherent limitation of \emph{MUSER}. Furthermore, \emph{ORCD} enhances interaction between headlines and reasoning pairs using three BERT encoders, surpassing \emph{SheepDog}’s multi-task framework. Compared to \emph{NRFE-D}, \emph{ORCD} constructs an efficient and interpretable pipeline by utilizing stance-conditioned generation in an unsupervised contrastive learning setup.  

Finally, we conduct an ablation study on ORCD to examine the impact of individual components on system performance (Section~\ref{sec:ablation}). Then, we empirically investigate the impact of the degree of sycophancy on outcomes (Section~\ref{sec:Sycophancy}) and conduct a cost-benefit analysis (Section~\ref{sec:cost_analysis}). Furthermore, an empirical study is also conducted to inspect the impact of the hyperparameters defined in our methodology (Section~\ref{sec:Hyper-Parameter}).

\subsection{Ablation Study}
\label{sec:ablation}
\begin{figure*}[]
    \centering
    \includegraphics[width=1\linewidth,height = 0.25\linewidth]{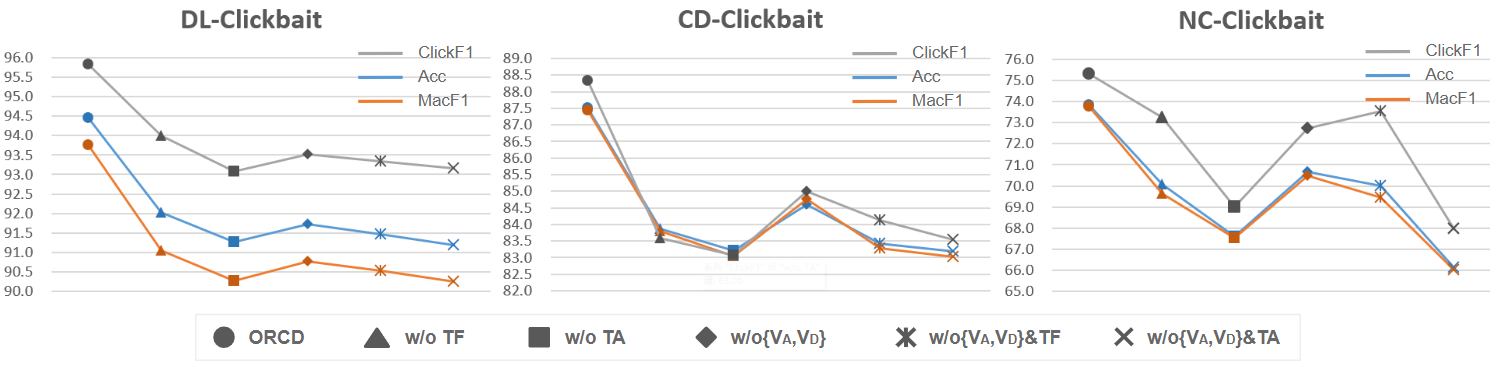}
    \caption{Ablation comparisons on three datasets among the five variants of \emph{ORCD} across three metrics. Specifically, w/o TF represents the original model that disables the title-free reasoning learner, w/o TA denotes that the title-aware reasoning learner is disabled in \emph{ORCD}, w/o $\{V_A, V_D\}$ means soft labels is frozen to use, w/o $\{V_A, V_D\}$\&TF implies that w/o TF and soft labels frozen are taken place simultaneously,  vice versa for w/o $\{V_A, V_D\}$\&TA.}
    \label{fig:ablation_3_1}
\end{figure*}

To inspect the impact of the model's components on affecting the performance of \emph{ORCD}, we conduct an ablation study by disabling each of the components in different variants of \emph{ORCD}. As shown in Fig.~\ref{fig:ablation_3_1}, the variants of our model include - w/o TF (a.k.a., disable the title-free reasoning learner), w/o TA (a.k.a., disable the title-aware reasoning learner), w/o $\{V_A, V_D\}$ (froze the utilization of soft labels), w/o $\{V_A, V_D\}$\&TF (disable the use of both title-free reasoning learner and soft labels) as well as w/o $\{V_A, V_D\}$\&TF. We have the following observations from the table.

\begin{table}[h]
\renewcommand{\arraystretch}{1.7}
\centering
\caption{Ablation study on the importance of sycophantic inference. Replacing sycophantic inference with label-consistent true inference (i.e., disabling polarity).}
\label{tab:no_syco}
\resizebox{1\linewidth}{!}{
\begin{tabular}{lccc}
\toprule
\textbf{} & \textbf{Accuracy (\%)} & \textbf{Macro-F1 (\%)} & \textbf{Clickbait-F1 (\%)} \\
\midrule
\textit{ORCD w/o Sycophantic Reasoning (¬ S.R.)} & 91.38 & 90.24 & 93.54 \\
\midrule
\textit{Performance Drops from Full ORCD} & \textbf{↓ 3.07} & \textbf{↓ 3.52} & \textbf{↓ 2.29} \\
\bottomrule
\end{tabular}
}
\end{table}

First, the absence of the Title-Free (w/o TF) or Title-Aware (w/o $T_A$) reasoning learners significantly reduces model performance across all metrics on both DLC and CDC datasets. This highlights the critical contribution of reasoning components in enhancing detection capabilities by providing semantic and contextual insights into the headlines. Second, the removal of soft labels (w/o {$V_A$, $V_D$}) has a noticeable negative effect on all metrics, especially on the CDC dataset. This emphasizes that the use of soft labels helps \emph{ORCD} adapt to nuanced and diverse clickbait styles, thereby improving generalizability and robustness. 

In addition, to further validate that the improvement of \emph{SORG} stems from leveraging sycophantic reasoning, rather than merely adding auxiliary reasoning, we replace flattering reasoning (beliefs that are consistent but not true) with label-consistent and factual reasoning on the DL-Clickbait dataset. As shown in Table~\ref{tab:no_syco}, removing sycophantic reasoning led to a consistent performance drop across all metrics, supporting our hypothesis that exploiting LLMs' alignment biases can provide effective contrastive signals.

\subsection{Sycophancy Level Analysis}
\label{sec:Sycophancy}

\begin{table}[htbp]
\renewcommand{\arraystretch}{1.7}
\setlength{\tabcolsep}{4pt}
\centering
\caption{As the reasoning progresses through multiple iterations, the sycophancy deepens.}
\label{tab:reasoning_ablation}
\resizebox{1\linewidth}{!}{
\begin{tabular}{c|cccc}
\hline
\textbf{Reasoning Iterations} & \textbf{Sycophancy Level} & \textbf{Acc} & \textbf{MacF1} & \textbf{ClickF1} \\
\hline
1st Round  & Low      & 91.57\% & 90.46\% & 93.70\% \\ \hline
2nd Round  & Moderate & 92.83\% & 91.88\% & 94.64\% \\  \hline
3rd Round  & High     & \textbf{93.61\%} & \textbf{92.67\%} & \textbf{95.30\%} \\
\hline
\end{tabular}
}
\end{table}

To examine how varying degrees of sycophancy affect experimental results, we conducted an investigation on the DL-Clickbait dataset by gradually increasing the number of inference iterations used to generate comparison pairs, resulting in more polarized inference content. The corresponding results are shown in the table~\ref{tab:reasoning_ablation}.

These results show a clear positive correlation between the level of sycophantic reasoning and downstream performance, reinforcing our claim that sycophancy-derived contrastive rationales provide meaningful learning signals for PLM-based models.

\subsection{Hyper-Parameter Settings}
\label{sec:Hyper-Parameter}
\begin{figure}[ht]
    \centering
    \includegraphics[width=0.98\linewidth,height = 0.65\linewidth]{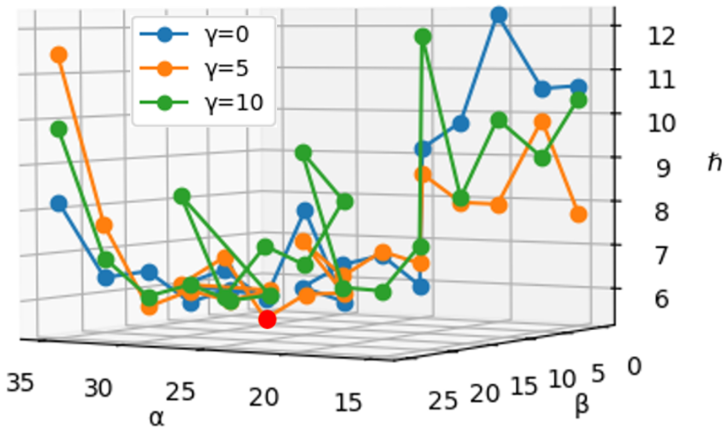}
    \caption{The alteration of the optimization objective $\hbar$ under the settings of different $\alpha$, $\beta$, and $\gamma$. The minimized $\hbar$ is marked in red color in the figure, which represents the best solution for \emph{SORG} conduction. }
    \label{fig:parameters}
\end{figure}
The hyper-parameters we inserted in \emph{SORG} include: (1) $\alpha$ (used in Algorithm 1
); (2) $\beta$ and $\gamma$ (used in Algorithm 2
). To optimize \emph{SORG}, we conduct an empirical study to select the best values of these parameters by considering the average time-consuming of reasoning generation (i.e., $T$) and the rating offset of reasoning-based titles (i.e., $P$), where polarity is the actual distance between the rating of the reasoning-based title (i.e., $V_A$ or $V_D$) and the value center (i.e., $50$). We expect to minimize $T$, yet maximize $P$ for tuning \emph{SORG}, thus the optimization objective is formalized as.
\begin{equation}
   {Minimize}\left\{ \hbar = \frac{T}{P} \right\}
\end{equation}
where $\hbar$ represents the macro average efficiency of the reasoning conduction process for news. Specifically, $T$ is the total consumed seconds for all data items, and $P$ is the total rating offset of the reasoning-based data. Thus, we aim to find the best $\alpha$, $\beta$, and $\gamma$ to minimize the value of $\hbar$. Fig.~\ref{fig:parameters} illustrates the fluctuation of $\hbar$ by varying the values of the three parameters. It can be observed from Fig.~\ref{fig:parameters} that $\hbar$ is minimized while the values of $\alpha$, $\beta$ and $\gamma$ are assigned with $30$, $10$, and $5$, respectively.

\section{Conclusions}
Unlike existing researches that attempt to avoid the appearance of LLM sycophancy, this study initially proposes to leverage the ability of LLM sycophancy to align with the prompts that embrace our preset opposing stances. To achieve this goal, we present a novel self-renewal opposing stance reasoning generation method (a.k.a., \emph{SORG}) to generate high-quality agree and disagree reasoning coupled with their ratings via LLM reflection. It is worth noting that \emph{SORG} is without using the label of news headlines, which allows its scalability and applicability to new domains. To utilize the generated opposing reasoning assisting clickbait title detection, we design a local task model named \emph{ORCD}, which contains three BERT Encoders to represent the news title and the corresponding agree and disagree reasoning for contrastive learning. In addition, the ratings of the reasoning pairs are also used as soft labels to tune the parameters of the title-free and title-aware reasoning learners. Our experimental results demonstrate the superiority of \emph{ORCD} by comparing it with three types of baselines. The ablation study reveals the importance of the two reasoning learners and soft labels. In our future work, the content of news will be considered as the reference for reasoning generation. Meantime, we will consider evaluating the availability of \emph{SORG} in other discriminative tasks, such as sentiment analysis, misinformation detection, and machine-generated text detection by taking advantage of the nature of the unsupervised opposing-stance reasoning mechanism in \emph{SORG}. 


\begin{acks}
This study is partially supported by the National Natural Science Foundation of China (62403412), the Natural Science Foundation of the Higher Education Institutions of Jiangsu Province of China under grant 23KJB520040, and the National Language Commission of China (ZDI145-71).
\end{acks}

\bibliographystyle{ACM-Reference-Format}
\balance
\bibliography{main}


\appendix

\section{Cost Analysis}
\label{sec:cost_analysis}

\begin{table}[h!]
\renewcommand{\arraystretch}{1.7}
\centering
\caption{Comparison of inference cost and clickbait detection performance on the DL-Clickbait (DLC) dataset.}
\label{tab:dlc_inference}
\resizebox{1\linewidth}{!}{
\begin{tabular}{l|cccc}
\hline
\textbf{Method} & \textbf{Time / Sample (s)} & \textbf{Input Tokens} & \textbf{Output Tokens} & \textbf{Clickbait-F1 (\%)} \\
\hline
GPT-4o (zero-shot)  & 0.80 & 198.28  & 142.57 & 75.96 \\  \hline
GPT-4o (few-shot)   & 2.30 & 2657.14 & 257.89 & 81.42 \\  \hline
ORCD (GPT-4o)       & 7.20 & 2501.60 & 309.43 & \textbf{95.83} \\  
\hline 
\end{tabular}
}
\end{table}

To evaluate the practical efficiency of different approaches, we measured the average inference time per sample, input/output token counts, and the resulting Clickbait-F1 on the DL-Clickbait dataset. As shown in Table~\ref{tab:dlc_inference}, our ORCD model achieves the best performance (95.83\% F1) but with greater inference cost, primarily due to multi-turn LLM generation and contrastive reasoning.

\tcbset{
  coltitle=black,                        
  fonttitle=\bfseries,                  
  colframe=blue!30!black,               
  colbacktitle=blue!25,                 
  colback=white,                        
  enhanced,
  boxrule=0.5pt,                        
  arc=1mm,                              
  sharp corners=south,                 
  boxsep=4pt,                           
  left=4pt, right=4pt, top=6pt, bottom=6pt,
  title filled=true,
  drop shadow southeast,                
  before skip=1pt,          
  after skip=20pt            
}

\section{Prompt Design}
\label{sec:prompt-design}

The following prompt template is used to instruct the LLM to assign a plausibility score (0–100) to a given headline.

\vspace{1em}

\begin{tcolorbox}[title=Prompt for Initial Title Rating]
\small
\textbf{Objective:} As news experts, rate the content of the headlines to assess how much people agree with the headlines.

\textbf{Requirement 1:} The content of the title is \{title\}.

\textbf{Requirement 2:} The score range is 0–100, where 0 means completely disagree, 50 means it is difficult to judge, and 100 means completely agree. The score should be humane.

\textbf{Requirement 3:} The output format is [int].
\end{tcolorbox}

To ensure that the initial reasoning aligns with a desired level of stance polarization, we design a follow-up prompt to trigger self-renewal in rating. This prompt instructs the LLM to reassess its original score , and either strengthen or weaken its stance based on a provided revision directive.

\vspace{1em}

\begin{tcolorbox}[title=Prompt for Self-Renewal Title Rating]
\small
\textbf{Objective:} Re-evaluate the consistency level based on the title content.

\textbf{Requirement 1:} The title content is \{title\}.

\textbf{Requirement 2:} Consider the consistency score of the previous title as \{previous score\}.

\textbf{Requirement 3:} The previous sense of identity does not meet the requirements, please \{increase/decrease\} your sense of identity.

\textbf{Requirement 4:} The score should be between 0 and 100.

\textbf{Requirement 5:} The output format is [int].
\end{tcolorbox}

To guide the LLM to generate structured agree or disagree reasoning for a given title, we design a prompt template that explicitly instructs the model to evaluate the headline from four aspects: common sense, logic, content completeness, and objectivity. The goal is to produce a short, persuasive justification that either supports or refutes the plausibility of the title.

\vspace{1em}

\begin{tcolorbox}[title=Prompt for Opposing-Stance Reasoning Generation]
\small

\textbf{Objective:} Make a comprehensive inference about the title from four aspects: common sense, logic, content integrity, and objectivity. The inference should make people \{agree/disagree\} the content in the title.

\textbf{Requirement 1:} The title content is \{title\}.

\textbf{Requirement 2:} Please agree with the title content in combination with the following four aspects:  
1. Common Sense: Does the title contain any information that is inconsistent with common sense or is incorrect?
2. Logic: Are there any leaps in reasoning or inconsistencies?
3. Content Completeness: Is there any information that is vague, intentionally left blank, or creates unnecessary suspense?
4. Objectivity: Is there any emotional manipulation or inflammatory language?

\textbf{Requirement 3:} The length of the reasoning should be limited to 40–60 words, and the content should be placed in {[ ]}.

\end{tcolorbox}

This question instructs LLMs to assign a new agreement score based on the title content, initial rating, and specific reasoning (agree or disagree). This score helps us verify whether the generated reasoning matches the expected position polarity.

\vspace{1em}

\begin{tcolorbox}[title=Prompt for Re-Scoring]
\small

\textbf{Goal:} Re-score based on the title content, initial score, and \{\textit{agree\_reason/disagree\_reason}\} reasoning.

\textbf{Requirement 1:} The title is \{title\}.

\textbf{Requirement 2:} The initial score is \{initial score\}.

\textbf{Requirement 3:} The reasoning content is \{agree\_reason/\\
disagree\_reason\}.

\textbf{Requirement 4:} The score should be between 0 and 100.

\textbf{Requirement 5:} The output format is {[int]}.

\end{tcolorbox}

This prompt instructs the LLM to analyze the given reasoning content from perspectives of rationality and logic, taking into account the previous reasoning and its associated score.

\vspace{1em}

\begin{tcolorbox}[title=Prompt for Reasoning Content Analysis]
\small

\textbf{Goal:} Analyze the \{\textit{reasoning type}\} reasoning content from the perspectives of rationality and logic.

\textbf{Requirement 1:} Consider the previous \{\textit{reasoning type}\} reasoning content: \{agree\_reason/disagree\_reason\}.

\textbf{Requirement 2:} Consider the previous score based on the \{\textit{reasoning type}\} reasoning: \{agree\_score/disagree\_score\}.

\textbf{Requirement 3:} The analysis should be limited to 50–70 words.

\textbf{Requirement 4:} Output format {[reasoning content]}.

\end{tcolorbox}

This prompt instructs the LLM to regenerate reasoning content because the previous reasoning failed to effectively influence the title's recognition score. The regeneration is guided by an analysis of logical inconsistencies and a comprehensive four-aspect evaluation to better align with the title content.

\vspace{1em}

\begin{tcolorbox}[title=Prompt for Regenerating Reasoning]
\small

\textbf{Goal:} Regenerate \{\textit{reasoning type}\} reasoning content, because the previous reasoning did not effectively \{\textit{effect verb}\} the title's recognition score.

\textbf{Requirement 1:} The title is \{title\}.

\textbf{Requirement 2:} The initial score is \{initial score\}.

\textbf{Requirement 3:} Consider the previous \{\textit{reasoning type}\} reasoning content: \{previous reasoning content\}.

\textbf{Requirement 4:} Consider the title score based on the previous \{\textit{reasoning type}\} reasoning: \{previous score\}.

\textbf{Requirement 5:} Consider the evaluation of the reasoning for \{\textit{reasoning type}\}: \{evaluation\}.

\textbf{Requirement 6:} Analyze the logical inconsistencies in the previous reasoning and explain why the new reasoning is more suitable for the title content.

\textbf{Requirement 7:} New inference generation should combine the following four aspects and adapt to the content of the title to \{\textit{effect verb}\} people's identification with the content of the title and make people \{\textit{emotion verb}\} in the content of the title.\\
1. Common Sense: Does it contain information that is inconsistent with common sense or is obviously wrong?\\
2. Logic: Are there any leaps in reasoning or inconsistencies?\\
3. Content Completeness: Is there any information that is vague, intentionally left blank, or creates unnecessary suspense?\\
4. Objectivity: Is there any judgement, emotional manipulation or inflammatory language?

\textbf{Requirement 8:} The limit for inference is 40–60 words, and the limit for explanation is 20–40 words. The inference content is placed in \texttt{[ ]} and the explanation content is placed in \texttt{( )}.

\textbf{Requirement 9:} Output format is {[Reasoning Content] (Explanatory Content)}.

\textbf{Requirement 10:} The score should still range from 0 to 100, and it should be more humanized, not restricted to multiples of 5.

\textbf{Requirement 11:} Output format for the score is {[int]}.

\end{tcolorbox}

\section{Manual Evaluation}
\label{sec:manual evaluation}

To further validate the quality and usability of the generated contrastive reasoning pairs, we conducted a human evaluation on 50 randomly sampled title–reasoning pairs. Each pair includes both an agree and a disagree rationale generated via the proposed ORCD prompting strategy. Three NLP graduate students independently rated each agree/disagree pair across four qualitative dimensions using a 5-point Likert scale (1 = very poor, 5 = excellent). The evaluated dimensions and their results are summarized in Table~\ref{tab:manual_eval_quality}.

Additionally, in 87\% of the evaluated cases, annotators agreed that the two rationales offered meaningful and interpretable contrast, further supporting the effectiveness of our approach in producing high-quality, instructive opposing reasoning.

\begin{table}[H]
\centering
\renewcommand{\arraystretch}{1.4}
\setlength{\tabcolsep}{2.5pt}
\caption{Human evaluation results for contrastive reasoning across four quality dimensions (1–5 scale).}
\label{tab:manual_eval_quality}
\begin{tabular}{p{2.5cm}cp{3.5cm}}
\toprule
\textbf{Dimension} & \textbf{Mean ± SD} & \textbf{Definition} \\
\midrule
\textbf{Fluency} & 4.6 ± 0.3 & Whether the rationale is grammatically correct, coherent, and natural in language usage. \\
\midrule
\textbf{Relevance} & 4.4 ± 0.4 & Whether the rationale is focused on and closely related to the specific content of the title. \\
\midrule
\textbf{Contrastiveness} & 4.2 ± 0.5 & Whether the agree/disagree rationales present clearly opposite and interpretable perspectives. \\
\midrule
\textbf{Informativeness} & 4.1 ± 0.5 & Whether each rationale provides unique, non-redundant insights into the clickbait nature of the title. \\
\bottomrule
\end{tabular}
\end{table}

\begin{table*}[h!]
\renewcommand{\arraystretch}{1.7}
\centering
\caption{Representative examples of clickbait and non-clickbait headlines along with their corresponding reasoning generated using our ORCD framework.}
\begin{tabular}{|p{1.2cm}|p{1.5cm}|p{4.3cm}|p{4.3cm}|p{4.3cm}|}  
\hline
\textbf{Label} & \textbf{Title} &   \textbf{Agree Reasoning (ORCD)} &   \textbf{Disagree Reasoning (ORCD)} &   \textbf{Direct Analysis (No ORCD)} \\ 
\hline
Non-clickbait & \parbox{1.5cm}{Tanzania accused of backtracking over sale of Masai’s ancestral land} & \parbox{4.3cm}{The title “Tanzania accused of backtracking over sale of Masai’s ancestral land” is consistent with common sense, as such issues often arise in disputes over indigenous lands. Logically, it follows that an accusation would arise if the government interfered with ancestral lands. The title is clear and objective without inflammatory language.} 
& \parbox{4.3cm}{The title presumes a deliberate "backtracking" by Tanzania without sufficient clarification, which contradicts common legal processes in land disputes. Additionally, there’s a logical leap without key context, leaving vague details open to misinterpretation or bias. This reasoning emphasizes how the title lacks crucial facts and could mislead or oversimplify.} 
& \parbox{4.3cm}{The headline uses emotionally charged words like “accused” and references a sensitive issue involving indigenous land, creating tension and curiosity. It lacks key details, prompting clicks to learn more. This makes it moderately clickbaity.} \\ 
\hline

\hline
Clickbait & \parbox{1.5cm}{Her Baby Is Asleep. Now Watch What The Puppy Does… Oh My God! (VIDEO)} & \parbox{4.3cm}{The title follows common sense by depicting a typical scenario involving a baby and a puppy, logical because pets often interact with babies in unexpected ways, and sufficiently complete, although it builds suspense. There is minimal emotional manipulation but without inflammatory language, retaining overall objectivity.} 
& \parbox{4.3cm}{The title lacks common sense by implying that a puppy’s action is shocking, which is unlikely. Logically, it leaps from mundane to extraordinary without reason. It's incomplete, offering suspense with no context. The phrase “Oh My God!” is emotional manipulation, exaggerating a likely trivial event.} 
& \parbox{4.3cm}{This headline is highly clickbaity. It uses vague phrasing ("what the puppy does") and exaggerated reaction cues ("Oh My God!") to trigger curiosity. The lack of concrete information and the emotional bait encourage users to click without knowing what to expect.} \\ 
\hline

\hline
Non-clickbait & \parbox{1.5cm}{Orlando Shooting A 'Very Serious Tragedy,' Says Dalai Lama} & \parbox{4.3cm}{The title reflects a common-sense understanding: the Orlando shooting was a highly publicized tragedy, and the Dalai Lama, a figure known for advocating peace, would likely identify it as such. The information is logical, clear, and conveys no emotional manipulation.} 
& \parbox{4.3cm}{The title potentially amplifies a mundane truth by attributing it to a spiritual leader, creating an unwarranted sense of novelty. Mass shootings are universally acknowledged as tragedies, and linking the Dalai Lama's statement doesn’t necessarily add new value. There’s a logical leap suggesting unnecessary validation, perhaps creating artificial importance.} 
& \parbox{4.3cm}{The headline is straightforward and informative. It reports a public figure’s reaction to a known event without using sensational or ambiguous language. There’s no attempt to provoke curiosity or exaggerate, so it’s not clickbait.} \\ 
\hline

\hline
Clickbait & \parbox{1.5cm}{	Study: Twenty-Something Women Love This Sort of Sexual Experimentation} & \parbox{4.3cm}{The title fits common sense, as people in their twenties often explore their identities, including sexuality. Logically, it makes sense that this age group might be curious about new experiences. The headline is complete, directly stating a study result. Objectively, it's neutral and avoids emotional manipulation or inflammatory language.} 
& \parbox{4.3cm}{The title makes a sweeping generalization, inconsistent with common sense, ignoring diversity in preferences and personalities. Logically, it rushes to conclusions about an entire age group without nuance. It lacks content depth, omitting study details or rigorous findings. The language appears subjective, bordering on sensationalism.} 
& \parbox{4.3cm}{This headline is clickbaity. It uses vague wording (“this sort of sexual experimentation”) to spark curiosity without revealing specifics, and phrases like “twenty-something women love” generalize a group to attract attention. The goal is clearly to provoke interest through suggestion and ambiguity.} \\ 
\hline

\hline
\end{tabular}

\label{tab:clickbait_examples}
\end{table*}

\section{Demos}
\label{sec:reasoning examples}

Table~\ref{tab:clickbait_examples} provides representative examples of the generated agree and disagree reasoning used in our ORCD (Opposing Reasoning-based Clickbait Detection) model. Each row corresponds to a real-world headline labeled as either clickbait or non-clickbait. The reasoning is generated by prompting an LLM under two opposing assumptions—agreeing or disagreeing with the title’s content—along four key dimensions: common sense, logic, completeness, and objectivity. A direct analysis is also provided for comparison, reflecting conventional LLM judgment without opposing-stance prompting. These examples illustrate how ORCD enhances interpretability and contrastive signal quality, enabling more robust detection of subtle clickbait cues.

\end{document}